\documentclass{article}
\usepackage{spconf,amsmath,graphicx,bm}
\usepackage{multirow}

\title{Improving RNN Transducer Modeling for End-to-End Speech Recognition}
\name{Jinyu Li, Rui Zhao, Hu Hu\sthanks{The work was done as an intern at Microsoft.}, and Yifan Gong}
\address{Speech and Language Group, Microsoft}
\begin{document}
\ninept
\maketitle
\begin{abstract}
In the last few years, an emerging trend in automatic speech recognition research is the study of end-to-end (E2E) systems.  Connectionist Temporal Classification (CTC), Attention Encoder-Decoder (AED), and RNN Transducer (RNN-T) are the most popular three methods. Among these three methods, RNN-T has the advantages to do online streaming which is challenging to AED and it doesn't have CTC's frame-independence assumption. In this paper, we improve the RNN-T training in two aspects. First, we optimize the training algorithm of RNN-T to reduce the memory consumption so that we can have larger training minibatch for faster training speed. Second, we propose better model structures so that we obtain  RNN-T models with the very good accuracy but small footprint. Trained with 30 thousand hours anonymized and transcribed Microsoft production data, the best RNN-T model with even smaller model size (216 Megabytes) achieves up-to 11.8\%  relative word error rate (WER) reduction from the baseline RNN-T model. This best RNN-T model is significantly better than the device hybrid model with similar size by achieving up-to 15.0\% relative WER reduction, and obtains similar WERs as the server hybrid model of 5120 Megabytes in size. 
\end{abstract}
\begin{keywords}
RNN Transducer, LSTM, GRU, layer trajectory, speech recognition
\end{keywords}
\section{Introduction}
\label{sec:intro}

Recent advances in automatic speech recognition (ASR) have been mostly due to the advent of using deep learning algorithms to build hybrid ASR systems with deep acoustic models like Deep Neural Networks (DNNs), Convolutional Neural Networks (CNNs), and Recurrent Neural Networks (RNNs). 
However, one problem with these hybrid systems is that there are several intermediate models (acoustic model, language model, lexicon model, decision tree, etc.) which either need expert linguistic knowledge or need to be trained separately. 
In the last few years, an emerging trend in ASR is the study of end-to-end (E2E) systems \cite{
sak2015learning, miao2015eesen, Chan-LAS, prabhavalkar2017comparison, battenberg2017exploring, sak2017recurrent, hadiantowards, chiu2018state, sainath2018improving, li2018advancing}. The E2E ASR system directly transduces an input sequence of acoustic features to an output sequence of tokens (phonemes, characters, words etc). This reconciles well with the notion that ASR is inherently a sequence-to-sequence task mapping input waveforms to output token sequences. Some widely used contemporary E2E approaches for sequence-to-sequence transduction are: (a) Connectionist Temporal Classification (CTC) \cite{Graves-CTCFirst, Graves-E2EASR}, (b) Attention Encoder-Decoder (AED) \cite{Cho-RNNEncDecSMT, Bahdanau-RNNEncDecAlignTranslate, bahdanau2016end, Chorowski-AttentionASR, Chan-LAS}, and (c) RNN Transducer (RNN-T)\cite{Graves-RNNSeqTransduction}. These approaches have been successfully applied to large-scale ASR systems \cite{sak2015learning, miao2015eesen, Chan-LAS, prabhavalkar2017comparison, battenberg2017exploring, chiu2018state, soltau2016neural, rao2017exploring, he2019streaming}.

Among all these three approaches, AED (or LAS: Listen, Attend and Spell \cite{Chan-LAS, chiu2018state}) is the most popular one. It contains three components: Encoder which is similar to a conventional acoustic model; Attender which works as an alignment model; Decoder which is analogous to a conventional language model. However, it is very challenging for AED to do online streaming, which is an important requirement for  ASR services. Although there are several studies towards that direction, such as monotonic chunkwise attention \cite{chiu2017monotonic} and triggered attention \cite{moritz2019triggered}, it's still a big challenge. As for CTC, it enjoys its simplicity by only using an encoder to map the speech signal to target labels. However, its frame-independence assumption is most criticized. There are several attempts to improve CTC modeling by relaxing or removing such assumption. In \cite{Das18CTCAttention, Das19}, attention modeling was directly integrated into the CTC framework to relax the frame-independence assumption by working on model hidden layers without changing the CTC objective function and training process, hence enjoying the simplicity of CTC training.  

A more elegant solution is RNN-T \cite{Graves-RNNSeqTransduction}, which extends CTC modeling by incorporating acoustic model with its encoder, language model with its prediction network, and decoding process with its joint network. There is no frame-independence assumption anymore and it is very natural to do online streaming. As a result, RNN-T instead of AED was successfully deployed to Google's device \cite{he2019streaming} with great impacts. In spite of its recent success in industry, there is less research in RNN-T when compared to the popular AED or CTC,  possibly due to the complexity of RNN-T training \cite{bagby2018efficient}. For example, the encoder and prediction network compose a grid of alignments, and the posteriors need to be calculated at each point in the grid to perform the forward-backward training of RNN-T. This three-dimension tensor requires much more memory than what is required in AED or CTC training. Given there are lots of network structure studies in hybrid systems (e.g., \cite{sainath2015convolutional, Li15FLSTM, Li16TFLSTM, sainath2016modeling, hsu2016prioritized,  zhao2016multidimensional, kim2017residual, HighwayBLSTM-zhang2016, pundak2017highway}), it is also desirable to explore advanced network structures so that we can put a RNN-T model into devices with both good accuracy and small footprint.

In this paper, we use the methods presented in the latest work \cite{he2019streaming} as the baseline and improve the RNN-T modeling in two aspects. First, we optimize the training algorithm of RNN-T to reduce the memory consumption so that we can have larger training minibatch for faster training speed. Second, we propose better model structures than the layer-normalized long short-term memory (LSTM) used in \cite{he2019streaming} so that we obtain  RNN-T models with better accuracy but smaller footprint. 

This paper is organized as follows. In Section \ref{sec:rnnt}, we briefly describe the basic RNN-T method. Then, in Section \ref{sec:training}, we propose how to improve the RNN-T training by reducing the memory consumption which constrains the training minibatch size because of the large-size tensors in RNN-T. In Section \ref{sec:model}, we propose several model structures which improve the baseline LSTM model used in  \cite{he2019streaming} for better accuracy and smaller model size. We evaluate the proposed models in Section \ref{sec:exp} by training them with 30 thousand hours anonymized and transcribed production data, and evaluating with several ASR tasks. Finally, we conclude the study in Section \ref{sec:con}.

\section{RNN-T}
\label{sec:rnnt}

\begin{figure}[t]
  \centering
  \includegraphics[width=0.6\linewidth]{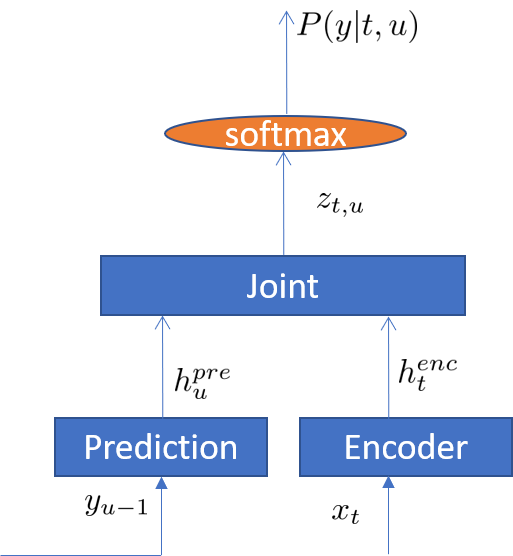}
  \caption{Diagram of RNN-Transducer.}
  \label{fig:RNNT}
\end{figure}

Figure \ref{fig:RNNT} shows the diagram of the RNN-T model, which consists of encoder, prediction, and joint networks. The encoder network is analogous to the acoustic model, which converts the acoustic feature $x_t$ into a high-level representation $h_t^{enc}$, where $t$ is time index.
\begin{equation}
h_t^{enc} = f^{enc} (x_t)
\end{equation}

The prediction network works like a RNN language model, which produces a high-level representation $h_u^{pre}$ by conditioning on the previous non-blank target $y_{u-1}$ predicted by the RNN-T model, where $u$ is output label index. 
\begin{equation}
h_u^{pre} = f^{pre} (y_{u-1})
\end{equation}

The joint network is a feed-forward network that combines the encoder network output $h_t^{enc}$ and the prediction network output $h_u^{pre}$ as 
\begin{align}
z_{t,u}  &= f^{joint} (h_t^{enc}, h_u^{pre}) \\
				&= \psi(U h_t^{enc} + V h_u^{pre} + b_z), \label{eq:z}
\end{align}
where $U$ and $V$ are weight matrices, $b_z$ is a bias vector, and $\psi$ is a non-linear function, e.g. Tanh or ReLU \cite{nair2010rectified}.

The $z_{t,u}$ is connected to the output layer with a linear transform
\begin{align}
h_{t,u}=W_y z_{t,u} +b_y.\label{eq:h}
\end{align}
The final posterior for each output token $k$ is obtained after applying softmax operation
\begin{align}
P(k|t,u)=softmax (h_{t,u}^k).
\label{eq:soft}
\end{align}

The loss function of RNN-T is the negative log posterior of output label sequence $\bf{y}$ given input acoustic feature $\bf{x}$,
\begin{align}
L = -ln P(\bf{y}|\bf{x}),
\end{align}
which is calculated based on the forward-backward algorithm described in \cite{Graves-RNNSeqTransduction}. 
The derivatives of loss $L$ with respect to $P⁡(k|t,u)$ is 
\begin{align}
\frac{\partial L} {\partial P⁡(k|t,u)}= -\frac {\alpha(t, u)} { P(\bf{y}|\bf{x})}\left\{ 
                \begin{array}{ll}
                  \beta(t, u+1) & \text{if} \quad k=y_{u+1}\\
                  \beta(t+1, u) & \text{if} \quad k=\emptyset \\
                  0 & \text{otherwise}
                \end{array}
								\right.
								\label{eq:partialP}
\end{align}
where $\emptyset$ denotes the blank label. $\alpha(t, u)$ is  is the probability of outputting ${\bf{y}}_{[1:u]}$ during ${\bf{x}}_{[1:t]}$ while $\beta(t,u)$ is the probability of outputting ${\bf{y}}_{[u+1:U]}$ during ${\bf{x}}_{[t:T]}$, assuming one sequence with acoustic feature length as T and label sequence length as U. 

\section{Training Improvement}
\label{sec:training}

We use  $K$ to denote the number of all output labels (including blank), and $D$ to denote the dimension of $z_{t,u}$. From Eq. \eqref{eq:z} and \eqref{eq:h},  we can see the size of $z$ is $T*U*D$, and the size of  $h$ is $K*T*U$. Compared with other E2E models, such as CTC or AED, RNN-T model consumes much more memory during training. This restricts the training minibatch size, especially when trained on GPUs with limited memory capacity. The training speed would be slow if small minibatch is used, therefore reducing memory cost is important for fast RNN-T training.  In this paper, we use two methods to reduce memory cost for RNN-T model training. 

\subsection{Efficient encoder and prediction output combination}
To improve the training efficiency of RNN, usually several sequences are used in parallel in one minibatch.  Given $N$ sequences in one minibatch with acoustic feature lengths $(T_1,T_2,...,T_N)$, the dimension of $h^{enc}$ would be $(N, max(T_1,T_2,...,T_N ),F)$ ($F$ is the size of vector $h_t^{enc}$) when they are paralleled in one minibatch. Because the actual element number in $h^{enc}$ is $\sum_{n=1}^N T_n *F$, some of the memory is wasted. Such waste is worse for RNN-T when combining encoder and prediction network output with the broadcasting method which is a popular implementation for dealing with different-size tensors in neural network training tools such as PyTorch. 

Suppose the label length of N sequences are $(U_1,U_2,...,U_N)$. The dimension of $h^{pre}$ would then be $(N,max(U_1,U_2,...,U_N ),F)$.  To combine $h^{enc}$ and $h^{pre}$ with the broadcasting method, the dimension of $h^{enc}$ and $h^{pre}$ is expanded to 
\begin{align}
(N, 1, max(T_1,T_2,...,T_N ), F)
\end{align}
and
\begin{align}
(N, max(U_1,U_2,...,U_N ), 1, F)
\end{align}
respectively. Then, they are combined according to Eq. \eqref{eq:z} and the output dimension becomes
\begin{align}
(N, max(U_1,U_2,...,U_N ),  max(T_1,T_2,...,T_N ), D).	\label{eq:joint}
\end{align}
Note the last dimension is $D$ instead of $F$ because of the projection matrices in Eq. \eqref{eq:z}. Hence, $z$ becomes a four-dimension tensor, which requires very large memory $(N* max(U_1,U_2,...,U_N )*  max(T_1,T_2,...,T_N )* D)$ to accommodate. 

To eliminate such memory waste, we did not use the broadcasting method to combine $h^{enc}$ and $h^{pre}$.  Instead, we implement the combination sequence by sequence. Hence, the size of $z_n$ for utterance $n$ is $T_n *U_n*D$ instead of  $max(T_1,T_2,…,T_N )*max⁡〖(U_1,U_2,…,U_N )*D$. Then we concatenate all $z_n$ instead of paralleling them, which means we convert $z$ into a two-dimension tensor $(\sum_{n=1}^N T_n *U_n), D)$.  In this way, the total memory cost for $z_1, z_2,...z_N$ is $(\sum_{n=1}^N T_n *U_n)*D$. This significantly reduces the memory cost, compared to the broadcasting method. There is no recurrent operation after the combination, hence such conversion will not affect the training speed for the  operations following it. Of course, we need to pass the sequence information like $T_n$, $U_n$  to any later operations so that they could process the sequences correctly based on such information.

Another way to reduce memory waste is to sort the sequences with respect to both acoustic feature and label length, and then performing the training with sorted sequences. However, from our experience, this generates worse accuracy than presenting randomized utterances to the training. In the deep speech 2 work \cite{amodei2016deep}, SortaGrad was used to present the sorted utterances to training in the first epoch for the better initialization of CTC training. After the first epoch, the training then is reverted back to a random order of data over minibatches, which indicates the importance of data shuffle.

\subsection{Function merging}

Most of the neural network training tools (e.g., PyTorch) are modular based. Although this is convenient for user to try various model structure, the memory usage is not efficient. For RNN-T model, if we define softmax and loss function with two separate modules, to get the derivative of loss with respect to $h_{t,u}^k$, i.e., $\frac{\partial L}{\partial h_{t,u}^k}$, we may use the chain rule 
\begin{align}
\frac{\partial L}{\partial h_{t,u}^k} = \frac{\partial L}{\partial P⁡(k|t,u)} \frac{\partial P⁡(k|t,u)}{\partial h_{t,u}^k} \label{eq:chain}
\end{align}
to first calculate $\frac{\partial L}{\partial P⁡(k|t,u)}$ with Eq. \eqref{eq:partialP} and then$\frac{\partial P⁡(k|t,u)}{\partial h_{t,u}^k}$. However, calculating and storing $\frac{\partial L}{\partial P(k|t,u)}$  is not necessary. Instead we directly derive the formulation of   $\frac{\partial L}{\partial h_{t,u}^k}$ as
\begin{align}
{\scriptsize 	
\frac{\partial L}{\partial h_{t,u}^k} 
=\frac {P⁡(k|t,u) \alpha(t, u)}{ P(\bf{y}|\bf{x})} \left[ \beta(t, u) - \left\{ 
                \begin{array}{ll}
                  \beta(t, u+1) & \text{if} \ k=y_{u+1}\\
                  \beta(t+1, u) & \text{if} \ k=\emptyset \\
                  0 & \text{otherwise}
                \end{array}
								\right. \right]	}.
\label{eq:der_h}
\end{align}
The tenor size of $\frac{\partial L}{\partial P(k|t,u)}$ of one minibatch is very big, hence  Eq. \eqref{eq:der_h} saves large memory compared with Eq. \eqref{eq:chain}. To further optimize this, we calculate softmax in-place \footnote{the operation that directly changes the content of a given tensor without making a copy.} for Eq. \eqref{eq:soft} and let $\frac{\partial L} {\partial h_{t,u}^k}$ take the memory storage place of $P⁡(k|t,u)$ in   Eq. \eqref{eq:der_h} . In this way, for this part  we only have one very large tensor in the memory (either  $P⁡(k|t,u)$ or $\frac{\partial L} {\partial h_{t,u}^k}$), compared to standard chain rule implementation which needs to store three large tensors ( $P⁡(k|t,u)$, $\frac{\partial L} {\partial h_{t,u}^k}$, and $\frac{\partial L}{\partial P(k|t,u)}$ ) in the memory. 

With above two methods, memory cost is reduced significantly, which helps to increase the training minibatch size. For a RNN-T model with around 4,000 output labels, the minibatch size could be increased from 2000 to 4000 when it is trained with V100 GPU which has 16 Gigabytes memory. The improvement is even larger for a RNN-T model with 36,000 output labels. The minibatch size could be increased from 500 to 2000. 

\section{Model Structure Exploration}
\label{sec:model}

In the latest successful RNN-T work \cite{he2019streaming}, the layer-normalized LSTMs with projection layer were used in both encoder and prediction networks. We describe it in Section \ref{ssec:LSTM} and use it as the baseline structure in this study. 
From Section \ref{ssec:ltLSTM} to Section \ref{ssec:cltGRU}, we propose different structures for RNN-T.  These new structures achieve more modeling power by 1) decoupling the target classification task and temporal modeling task with separate modeling units; 2) exploring future context frames to generate more informative encoder output. 

\subsection{Layer normalized LSTM}
\label{ssec:LSTM}
In \cite{he2019streaming}, layer normalization and projection layer for LSTM were reported important to the success of RNN-T modeling. Following \cite{lei2016layer}, we define the layer normalization function for vector $v$ given adaptive gain $\alpha$ and bias $\beta$ as 
\begin{align}
LN(v) &= (v- \mu)/\sigma \odot \alpha + \beta, \\
\mu &= \frac{1}{D_v} \sum_{i=1}^{D_v} v_i, \\
\sigma &= \sqrt{\frac{1}{D_v} \sum_{i=1}^{D_v} (v_i-\mu)^2}
\end{align}
where ${D_v}$ is the dimension of  $v$, and $\odot$ is element-wise product.

Then, we define the layer normalized LSTM function with project layer   ${h}_t^l = LSTM({h}_{t-1}^l, {x}_{t}^l, {c}_{t-1}^l)$ as  
  \begin{align}
  {i}_t^l &= \sigma ( LN({W}^l_{ix} {x}_{t}^l + {W}^l_{ih} {h}_{t-1}^l + {b}^l_{i})) \label{eq:igate}\\
  {f}_t^l &= \sigma ( LN({W}^l_{fx} {x}_{t}^l + {W}^l_{fh} {h}_{t-1}^l + {b}^l_{f})) \label{eq:fgate}\\
  {c}_t^l &= {f}_t^l \odot {c}_{t-1}^l + {i}_t^l \odot \phi( LN({W}^l_{cx} {x}_{t}^l + {W}^l_{ch} {h}_{t-1}^l + {b}^l_{c})) \label{eq:cell} \\
  {o}_t^l &= \sigma ( LN({W}^l_{ox} {x}_{t}^l + {W}^l_{oh} {h}_{t-1}^l  + {b}^l_{o})) \label{eq:ogate} \\
  {q}_t^l &= {o}_{t}^l \odot \phi(LN({c}_t^l)) \\ 
	{h}_t^l &= W_p^l q_t^l \label{eq:out}
  \end{align}
The vectors  ${i}_t^l$, ${o}_t^l$, ${f}_t^l$, ${c}_t^l$ are the activations of the input, output, forget gates, and memory cells, respectively. ${h}_t^l$ is the output of the LSTM.  ${W}^l_{.x}$ and  ${W}^l_{.h}$ are the weight matrices for the inputs ${x}_{t}^l$ and the recurrent inputs ${h}_{t-1}^l$, respectively. ${b}^l_{.}$ are bias vectors. The functions $\sigma$ and $\phi$ are the logistic sigmoid and hyperbolic tangent nonlinearity, respectively. 

For the multi-layer LSTM, $x_t^l = h_t^{l-1}$, then we have the hidden output of the $l$th layer at time $t$ as
\begin{align}
{h}_t^l &= LSTM(h_{t-1}^l, h_t^{l-1}, c_{t-1}^l),
\label{eq:LSTM}
\end{align}
and we use the last hidden layer output $h_t^L$ and $h_u^M$ of the encoder and prediction networks as $h_t^{enc}$ and $h_u^{pre}$, where $L$ and $M$ denote the number of layers in encoder and prediction networks respectively.

\subsection{Layer trajectory LSTM}
\label{ssec:ltLSTM}
\begin{figure}[t]
  \centering
  \includegraphics[width=\linewidth]{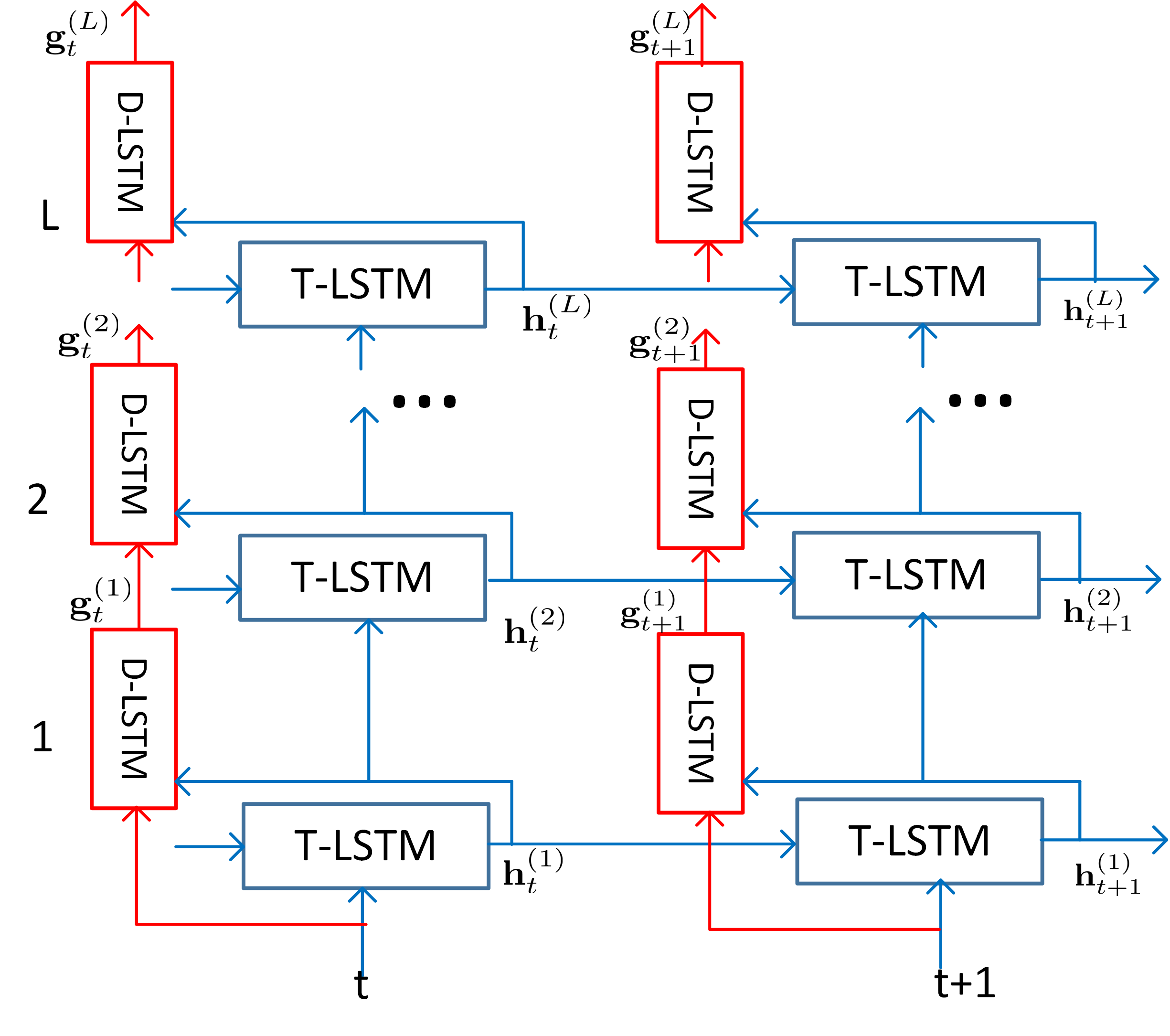}
  \caption{Diagram of layer trajectory LSTM (ltLSTM). 
	Depth-LSTM (D-LSTM) is used to scan the outputs of time-LSTM (T-LSTM) across all layers at the current time step to get summarized layer trajectory information for senone classification. 
	Note that There is no time recurrence in D-LSTMs, which only occurs in T-LSTMs.}
  \label{fig:ltLSTM}
\end{figure}

Recently, significant accuracy improvement was reported with a layer trajectory LSTM (ltLSTM) model \cite{Li18ltLSTM, Li18ltLSTMExplore} in which depth-LSTMs are used to scan the outputs of multi-layer time-LSTMs to extract information for label classification. By decoupling the temporal modeling and senone classification tasks, ltLSTM allows time and depth LSTMs focusing on their individual tasks. The model structure is illustrated by Figure \ref{fig:ltLSTM}. 

Following \cite{Li18ltLSTM}, we define the layer-normalized ltLSTM formulation using 
\begin{align}
{h}_t^l &= LSTM(h_{t-1}^l, h_t^{l-1}, c_{t-1}^l)	\label{eq:temporal} \\
{g}_t^l &= LSTM(h_{t}^l, {g}_t^{l-1}, d_t^{l-1}),	\label{eq:depth}
\end{align}
with the LSTM function defined from Eq. \eqref{eq:igate} to Eq. \eqref{eq:LSTM}, where $c_{t-1}^l$ and $d_t^{l-1}$ are the memory cells from time-LSTM (time $t-1$ and layer $l$) and depth-LSTM (time $t$ and layer $l-1$), respectively.  The time-LSTM formulation in Eq. \eqref{eq:temporal} performs temporal modeling, while the depth-LSTM formulation in Eq. \eqref{eq:depth} works across layers for the target classification. ${g}_t^l$ is the output of the depth-LSTM at time $t$ and layer $l$. Similarly, we can use the last hidden layer output $g_t^L$ and $g_u^M$ of the encoder and prediction networks as $h_t^{enc}$ and $h_u^{pre}$ for RNN-T.

\subsection{Contextual layer trajectory LSTM}
\label{ssec:cltLSTM}
In \cite{Li19cltLSTM}, the contextual layer trajectory LSTM (cltLSTM) was proposed to further improve the performance of ltLSTM models  by using context frames to capture future information. 
In cltLSTM,  ${g}_{t}^{l-1}$ used in Eq. \eqref{eq:depth} is replaced by the lookahead embedding vector ${\zeta}_{t}^{l-1}$ in order to incorporate the future context information.  The embedding vector is computed from the depth-LSTM using Eq. \eqref{eq:cltLSTM}
\begin{align}
{\zeta}_{t}^{l-1} &= \sum_{\delta =0}^{\tau}  {G}_{\delta}^{l-1} {g}_{t+\delta}^{l-1} \label{eq:cltLSTM} \\
{h}_t^l &= LSTM(h_{t-1}^l, h_t^{l-1}, c_{t-1}^l)	\\
{g}_t^l &= LSTM(h_{t}^l, \bm{\zeta}_{t}^{l-1}, d_t^{l-1}),	
\end{align}
where ${G}_{\delta}^{l-1}$ denotes the weight matrix applied to depth-LSTM output ${g}_{t+\delta}^{l-1}$. Note it is only meaningful to have future context lookahead for the encoder network. Therefore,  ${\zeta}_t^L$ of the encoder  network can be used as $h_t^{enc}$ for RNN-T. In Eq. \eqref{eq:cltLSTM}, every layer has $\tau$ frames lookahead. To generate ${\zeta}_t^L$, we have $L*\tau$ frames lookahead in total.

\subsection{Layer normalized gated recurrent units}
\label{ssec:GRU}
Similar to \cite{he2019streaming}, our target is to deploy RNN-T  for device ASR. Therefore, footprint is a major factor when developing models. As gated recurrent unit (GRU) \cite{chung2014empirical} usually has light weight compared to LSTM, we use it as the building block for RNN-T and define the layer normalized GRU function  $GRU({h}_{t-1}^l, {x}_{t}^l)$  as  
  \begin{align}
  {z}_t^l &= \sigma ( LN({W}^l_{zx} {x}_{t}^l + {W}^l_{zh} {h}_{t-1}^l + {b}^l_{z})) \\
  {r}_t^l &= \sigma ( LN({W}^l_{rx} {x}_{t}^l + {W}^l_{rh} {h}_{t-1}^l + {b}^l_{r})) \\
  {\tilde{h}}_t^l &= \phi( LN({W}^l_{hx} {x}_{t}^l + {W}^l_{hh} ({r}_t^l \odot {h}_{t-1}^l) + {b}^l_{h}))\\
  {h}_t^l &= {z}_{t}^l \odot {h}_{t-1}^l + (1-{z}_{t}^l) \odot {\tilde{h}}_t^l
  \end{align}
Again, ${W}^l_{.x}$ and  ${W}^l_{.h}$ are the weight matrices for the inputs ${x}_{t}^l$ and the recurrent inputs ${h}_{t-1}^l$, respectively. ${b}^l_{.}$ are bias vectors. 
For the multi-layer GRU, $x_t^l = h_t^{l-1}$, then we have the hidden output of the $l$th layer at time $t$ as
\begin{align}
{h}_t^l &= GRU(h_{t-1}^l, h_t^{l-1})	
\end{align}
Note that all the networks in this study are uni-directional and our layer normalized GRU is different from the bi-directional GRU with batch normalization used in \cite{battenberg2017exploring}. Layer normalization is better than batch normalization for recurrent neural networks \cite{lei2016layer}. 

\subsection{Layer trajectory GRU}
\label{ssec:ltGRU}
Similar to the layer normalized ltLSTM presented in Section \ref{ssec:ltLSTM}, we propose layer trajectory GRU (ltGRU) with layer normalization in this study as
\begin{align}
{h}_t^l &= GRU(h_{t-1}^l, h_t^{l-1})	\label{eq:t-gru} \\
{g}_t^l &= GRU(h_{t}^l, {g}_t^{l-1})	\label{eq:d-gru}. 
\end{align}
The time-GRU formulation in Eq. \eqref{eq:t-gru} performs temporal modeling, while the depth-GRU formulation in Eq. \eqref{eq:d-gru} works across layer for target classification. ${g}_t^l$ is the output of the depth-GRU at time $t$ and layer $l$. We can use the last hidden layer output $g_t^L$ and $g_u^M$ of the encoder and prediction networks as $h_t^{enc}$ and $h_u^{pre}$ for RNN-T.

\subsection{Contextual layer trajectory GRU}
\label{ssec:cltGRU}
To further improve the performance of ltGRU models, we extend ltGRU in Section \ref{ssec:ltGRU} by utilizing the lookahead embedding vector in order to incorporate the future context information.  The embedding vector ${\zeta}_{t}^{l-1}$ is computed from the depth-GRU using Eq. \eqref{eq:cltGRU}
\begin{align}
{\zeta}_{t}^{l-1} &= \sum_{\delta =0}^{\tau}  {q}_{\delta}^{l-1} \odot {g}_{t+\delta}^{l-1} \label{eq:cltGRU} \\
{h}_t^l &= GRU(h_{t-1}^l, h_t^{l-1})	\\
{g}_t^l &= GRU(h_{t}^l, \bm{\zeta}_{t}^{l-1})	
\end{align}
Different from Eq. \eqref{eq:cltLSTM}, Eq. \eqref{eq:cltGRU} uses vector ${q}_{\delta}^{l-1}$ instead of matrix ${G}_{\delta}^{l-1}$ to integrate future context ${g}_{t+\delta}^{l-1}$ from the depth-GRU. This further reduces the model footprint. Hence  we call this model as elementwise contextual layer trajectory GRU (ecltGRU). 

${\zeta}_t^L$  of the encoder  network can then be used as $h_t^{enc}$ for RNN-T.  In Eq. \eqref{eq:cltGRU}, every layer has $\tau$ frames lookahead, and we have $L*\tau$ frames lookahead in total to generate ${\zeta}_t^L$.

\section{Experiments}
\label{sec:exp}

In this section, we evaluate the effectiveness of the proposed models. In our experiments, all models are  uni-directional, and were trained with 30 thousand (k) hours of anonymized and transcribed Microsoft production data, including Cortana and  Conversation data, recorded in both close-talk and far-field conditions. We evaluated all models with  Cortana and Conversation test sets. Both sets contain mixed close-talk and far-field recordings, with 439k and 111k words, respectively. The Cortana  set has shorter utterances related to voice search and commands, while the Conversation  set has longer utterances from conversations. We also evaluate the models on the third test set named as DMA with 29k words, which is not in Cortana or Conversation domain. The DMA domain was unseen during the model training, and serves to evaluate the generalization capacity of the model. 

\subsection{RNN-T models with greedy search}

\begin{table}[t]
  \caption{WERs and sizes of all RNN-T models on Cortana, Conversation (Conv.), and DMA test sets. All test sets are mixed with close-talk and far-field recordings. All encoder (enc.) networks have 6 hidden layers, and all prediction (pred.) networks have 2 hidden layers. The decoding results are generated by greedy search. The cltLSTM and ecltGRU set $\tau=4$, i.e., using 4 future context frames at each layer.}
  \label{tab:wer}
  \centering
  \begin{tabular}{l|l|c|c|c|c}
    	\hline
							enc. & pred. & size Mb & 	Cortana \%				 & 			Conv. \%     & DMA \%       \\							
    	\hline
		LSTM & LSTM 		&	255 & 12.03 & 22.18 & 26.78\\
		ltLSTM & LSTM 		&	422 & 11.11 & 22.41 & 24.70\\
		ltLSTM & ltLSTM 		&	482 & 10.91 & 22.51 & 25.15\\
		cltLSTM & LSTM 		&	469 & 9.92 & 19.86 & 23.03\\
\hline
		GRU & GRU 		&	139 & 12.30 & 22.90 & 26.50\\
		ltGRU & GRU 		&	216 & 11.60 & 21.74 & 24.99\\
		ltGRU & ltGRU 		&	235 & 11.58 & 21.69 & 26.00\\
		ecltGRU & GRU 		&	216 & 10.68 & 19.76 & 23.22\\
    	\hline
  \end{tabular}
\end{table}

In Table \ref{tab:wer}, we compare different structures of RNN-T models in terms of size and word error rate (WER) using greedy search. The computational cost of RNN-T models during runtime is proportional to the model size. The feature is 80-dimension log Mel filter bank for every 10 milliseconds (ms) speech. Three of them are stacked together to form a frame of 240-dimension input acoustic feature to the encoder network. All encoder (enc.) networks have 6 hidden layers, and all prediction (pred.) networks have 2 hidden layers. The joint network always outputs a vector with dimension 640. The output layer models 4096 word piece units together with blank label. The word piece units are generated by running byte pair encoding \cite{sennrich2015neural} on the acoustic training texts. Similar to the latest successful RNN-T model on device \cite{he2019streaming}, our first model uses LSTM for both encoder and prediction networks. The encoder network has a 6 layer layer-normalized LSTM with 1280 hidden units at each layer and the output size then is reduced to 640 using a linear projection layer. We denote this layer-normalized LSTM structure as 1280p640. The prediction network has 2 hidden layers, and each layer is with the 1280p640 LSTM structure. This model has total size of 255 megabytes (Mb). We use this RNN-T model as the baseline model. 

Then, we explore ltLSTM structures described in Section \ref{ssec:ltLSTM} for encoder and prediction networks. All the LSTM components in ltLSTM use the 1280p640 LSTM structure. Simply using ltLSTM in the encoder significantly reduces the WERs on Cortana and DMA test sets, but increases the model size from 255 Mb to 422 Mb. Further using ltLSTM in prediction network increases the model size to 482 Mb, without benefiting the WER clearly. Hence, using ltLSTM only in the encoder is a good setup. 

Using cltLSTM described in Section \ref{ssec:cltLSTM} in the encoder network with $\tau=4$ at each layer improves the WERs significantly, with respectively 17.5\%, 10.5\%, and 14.0\% relative WER reduction on Cortana, Conversation, and DMA test sets  from the baseline  RNN-T model which uses layer-normalized LSTM in both encoder and decoder networks. This clearly shows the advantage of using future context for the encoder network. However, this model also has a very large model size as 469 Mb, which especially brings challenges to the deployment into devices. 

Next, layer-normalized LSTM is replaced with layer-normalized GRU described in Section \ref{ssec:GRU}. The GRU in this study is with 800 units at each layer without any projection layer. When using layer-normalized GRU in both encoder and prediction networks, we significantly reduce the RNN-T model size to 139 Mb and obtain slightly worse WERs than the baseline RNN-T model using LSTM. 

The RNN-T model using ltGRU proposed in Section \ref{ssec:ltGRU} in the encoder network has the model size 216 Mb, and significantly improves the RNN-T model with GRU in the encoder network. Again, applying ltGRU to prediction network doesn't bring any additional benefits but increases the model size. 

Finally, ecltGRU which uses $\tau=4$ at each layer has almost the same size as ltGRU in the encoder network due to the use of vectors instead of matrices when incorporating future context, but significantly improves the WERs because of the future context access.  The size of this model (216 Mb) is smaller than that (255 Mb) of the baseline RNN-T model which uses LSTM in the encoder and prediction networks. At the same time, it outperforms the baseline RNN-T model with  11.2\%, 10.9\% and 13.3\% relative WER reduction on Cortana, Conversation, and DMA test sets, respectively. 

\begin{figure}[t]
  \centering
  \includegraphics[width=\linewidth]{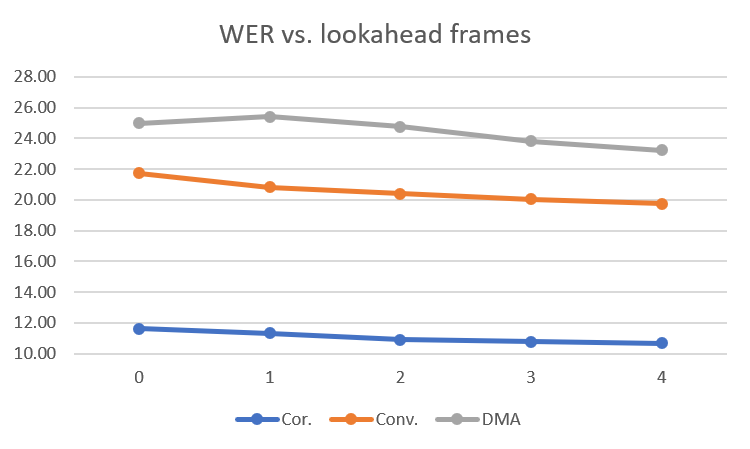}
  \caption{The WERs of the ecltGRU model with respect to $\tau$ future context frames  at each layer.}
  \label{fig:lookahead}
\end{figure}

Given the good performance of ecltGRU, we evaluate the impact of future frame contexts at each layer in Figure \ref{fig:lookahead} by varying the value of $\tau$. When $\tau=0$, the ecltGRU model just reduces to the ltGRU model. With larger $\tau$ value, the WERs drop monotonically. Setting $\tau$ as 3 or 4 doesn't have too much WER difference while smaller $\tau$ value does reduce the model latency with less future context access.

\subsection{Comparison with hybrid models}

\begin{table}[t]
  \caption{Comparison of hybrid models with RNN-T models on Cortana, Conversation (Conv.), and DMA test sets. The RNN-T decoding results are generated by beam search with beam width 10. The ecltGRU sets $\tau=4$, i.e., using 4 future context frames  at each layer.}
  \label{tab:comp}
  \centering
\begin{tabular}{l|l|l|c|c|c|c}
\hline
                        & \begin{tabular}[c]{@{}l@{}}AM \\ / enc.\end{tabular} & \begin{tabular}[c]{@{}l@{}}LM \\ / pred.\end{tabular}  & \begin{tabular}[c]{@{}c@{}}size \\ Mb\end{tabular} & \begin{tabular}[c]{@{}c@{}}Cortana \\ \%\end{tabular} & \begin{tabular}[c]{@{}c@{}}Conv. \\ \%\end{tabular} & \begin{tabular}[c]{@{}c@{}}DMA \\ \%\end{tabular} \\ \hline
\multirow{2}{*}{hybrid} & LSTM                                                 & 5gram                                                  & 5120                                               & 9.35                                               & 18.82                                               & 20.18                                             \\ \cline{2-7} 
                        & LSTM                                                 & \begin{tabular}[c]{@{}l@{}}pruned\\ 5gram\end{tabular} & 218                                                & 10.92                                              & 20.22                                               & 23.05                                             \\ \hline
\multirow{2}{*}{RNN-T}  & LSTM                                                 & LSTM                                                   & 255                                                & 9.94                                               & 19.70                                               & 23.19                                             \\ \cline{2-7} 
                        & ecltGRU                                              & GRU                                                    & 216                                                & 9.28                                               & 18.22                                               & 20.45                                             \\ \hline
\end{tabular}
\end{table}

In Table \ref{tab:comp}, we compare  RNN-T models decoded using beam search with hybrid models trained with cross entropy criterion on Cortana, Conversation, and DMA test sets. Note that all models, not matter RNN-T or hybrid, can be improved by sequence discriminative training.  The acoustic model of this hybrid model is a 6 layer LSTM with 1024 hidden units at each layer and the output size then is reduced to 512 using a linear projection layer. The softmax layer has 9404-dimension output to model the senone labels. 
The input feature is also 80-dimension log Mel filter bank for every 10 milliseconds (ms) speech. We applied frame skipping by a factor of 2 \cite{Miao16} to reduce the runtime cost, which corresponds to 20ms per frame. This acoustic model (AM) has the size of 120 Mb. The language model (LM) is a 5-gram with around 100 million (M) ngrams, which is compiled into a graph of 5 gigabytes (Gb) size. We refer this configuration as the server setup. We also prune the LM for the purpose of device usage, resulting in a graph of 98 Mb size. The decoding of these two hybrid models uses our production setup, differing only the LM size. We list the WERs of this AM combined with both the large LM and pruned LM in Table \ref{tab:comp}. Clearly, the device setup with small LM gives worse WERs than the server setup using  large LM. 

The RNN-T decoding results are generated by beam search with beam width 10. Comparing the results in Table \ref{tab:comp} and Table \ref{tab:wer}, we can see that beam search significantly improves the WERs from the greedy search. Using ecltGRU in the encoder network and GRU in the prediction network for RNN-T outperforms the baseline RNN-T which uses LSTM everywhere, with 6.6\%, 7.5\%, and 11.8\% relative WER reduction on  Cortana, Conversation, and DMA test sets, respectively. The size of this best RNN-T model is 216 Mb, less than the baseline RNN-T model with LSTM units.   Moreover, the WERs of this best RNN-T model matches the WERs of hybrid model with server setup which has a 5 Gb LM. Comparing to the device hybrid setup which has similar size (218 Mb), this best RNN-T obtains 15.0\%, 9.9\%, and 11.3\% relative WER reduction on  Cortana, Conversation, and DMA test sets, respectively. With the high recognition accuracy and small footprint, this RNN-T model is ideal for device ASR. 

\begin{figure}[t]
  \centering
  \includegraphics[width=\linewidth]{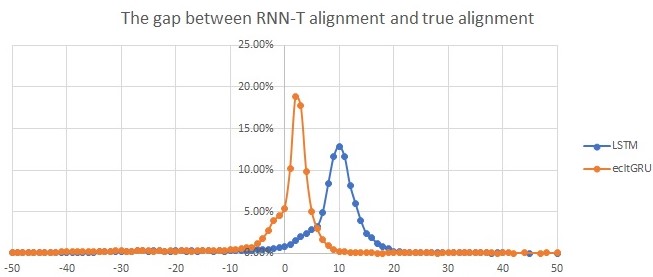}
  \caption{The gap between ground truth word alignment and the word alignment from two RNN-T models in Table \ref{tab:comp}. The ecltGRU sets $\tau=4$, i.e., using 4 future context frames  at each layer.}
  \label{fig:align}
\end{figure}

Finally, we look at the gap between ground truth word alignment obtained by force alignment with a hybrid model and the word alignment generated by greedy decoding from two RNN-T models in Table \ref{tab:comp}. As shown in Figure \ref{fig:align}, the baseline RNN-T with LSTM in the encoder network has larger delay than the ground truth alignment. The average delay is about 10 input frames. In contrast, the RNN-T model with ecltGRU has less alignment discrepancy, with average 2 input frames delay. This is because the ecltGRU encoder has total 24 frames lookahead, which provides more information to RNN-T so that it makes decision earlier than the baseline model. Because the input to RNN-T models spans 30ms by stacking 3 10ms frames together, therefore the average latency of RNN-T with ecltGRU encoder is (24+2)*30ms = 780ms while the average latency of RNN-T with LSTM encoder is 10*30ms = 300ms. Hence, partial accuracy advantage of ecltGRU encoder comes from the tradeoff of latency.

\section{Conclusions}
\label{sec:con}

In this paper, we improve the RNN-T training for end-to-end ASR from two aspects. First, we reduce the memory consumption of RNN-T training with efficient combination of the encoder and prediction network outputs, and by reformulating the gradient calculation to avoid storing multiple large tensors in the memory. In this way, we significantly increase the minibatch size (otherwise constrained by the memory usage) during training. Second, we improve the baseline RNN-T model structure which uses LSTM units by proposing several new structures. All the proposed structures use the concept of layer trajectory which decouples the classification task and temporal modeling task by using depth LSTM / GRU units and time LSTM / GRU units, respectively. The best tradeoff between model size and accuracy is obtained by the RNN-T model with ecltGRU in the encoder network and GRU in the prediction network. The future context lookahead at each layer helps to build accurate models with better performance. 

Trained with 30 thousand hours anonymized and transcribed Microsoft production data, this best RNN-T model achieves respectively 6.6\%, 7.5\%, and 11.8\% relative WER reduction on  Cortana, Conversation, and DMA test sets, from the baseline RNN-T model but with smaller model size (216 Megabytes), which is ideal for the deployment to devices. This best RNN-T model is significantly better than the hybrid model with similar size, by reducing 15.0\%, 9.9\%, and 11.3\% relative WER on  Cortana, Conversation, and DMA test sets, respectively. It also obtains similar WERs as the server-size hybrid model of 5120 Megabytes in size. 

\bibliographystyle{IEEEbib}
\bibliography{refs}

\end{document}